\documentclass[journal,twoside,web]{ieeecolor}
\usepackage{pifont}
\usepackage{xcolor}
\newcommand{\cmark}{\textcolor{green!60!black}{\ding{51}}}
\newcommand{\xmark}{\textcolor{red}{\ding{55}}}
\usepackage{generic}
\usepackage{cite}
\usepackage{amsmath,amssymb,amsfonts}
\usepackage{algorithmic}
\usepackage{graphicx}
\usepackage{algorithm,algorithmic}
\usepackage{hyperref}
\usepackage{booktabs}
\usepackage{threeparttable}
\hypersetup{hidelinks=true}
\usepackage{textcomp}
\usepackage{tabularx}
\newcolumntype{C}{>{\centering\arraybackslash}X}
\def\BibTeX{{\rm B\kern-.05em{\sc i\kern-.025em b}\kern-.08em
    T\kern-.1667em\lower.7ex\hbox{E}\kern-.125emX}}
\begin{document}
\title{RetiBridge: Bridging Quantitative Retinal Biomarkers and Qualitative Diagnosis with a Knowledge-Guided Multimodal Large Language Model}

\author{
Zhuangzhi Gao\textsuperscript{$\dagger$},
Hongyi Qin\textsuperscript{$\dagger$},
He Zhao,
Qinkai Yu,
Feixiang Zhou,
Fu Wang,
Jinru Ding,
Eduard Shantsila,
Uazman Alam,
Alena Shantsila,
Wahbi El-Bouri,
Gregory Y. H. Lip,
and Yalin Zheng
\thanks{\textsuperscript{$\dagger$}These authors contributed equally to this work.}
\thanks{Gregory Y. H. Lip, Alena Shantsila, Qinkai Yu and Wahbi El-Bouri are with the Liverpool Centre for Cardiovascular Science, University of Liverpool, Liverpool, United Kingdom.}
\thanks{Yalin Zheng, He Zhao, Feixiang Zhou, Fu Wang and Zhuangzhi Gao are with the Department of Eye and Vision Sciences, University of Liverpool, Liverpool, United Kingdom.}
\thanks{Zhuangzhi Gao and Jinru Ding are with the Shanghai Artificial Intelligence Laboratory, Shanghai, China.}
\thanks{Hongyi Qin is with 
Institute of Life Course \& Medical Sciences, University of Liverpool, Liverpool, United Kingdom}
\thanks{Alena Shantsila, Uazman Alam and Wahbi El-Bouri are Cardiovascular \& Metabolic Medicine, University of Liverpool, Liverpool, United Kingdom}
\thanks{Zhuangzhi Gao and Eduard Shantsila are the Department of Primary Care and Mental Health, University of Liverpool, Liverpool, United Kingdom}
\thanks{\raggedright Corresponding author: Yalin~Zheng (e-mail: \href{mailto:Yalin.Zheng@liverpool.ac.uk}{Yalin.Zheng@liverpool.ac.uk}).}
}
\maketitle

\begin{abstract}
Retinal biomarkers captured by color fundus photography and optical coherence tomography provide clinically valuable evidence for both ocular and systemic diseases. Multimodal large language models (MLLMs) have shown promise for retinal image interpretation, yet existing ophthalmic models rarely quantify these clinically relevant biomarkers or explicitly translate their measurements into qualitative, evidence-grounded diagnostic conclusions. To address this gap, we introduce RetiBridge, a knowledge-guided multimodal large language model that jointly analyzes color fundus photography (CFP), optical coherence tomography (OCT), and text, explicitly bridging quantitative retinal biomarkers to qualitative clinical sub-inferences and coherent diagnostic conclusions. RetiBridge combines knowledge-guided instruction generation, OCT–biomarker alignment, and supervised multimodal instruction tuning to learn a biomarker-grounded quantitative-to-qualitative diagnostic pathway. Using 15,611 paired CFP–OCT samples from UK Biobank with 31 OCT and 6 CFP biomarkers, we construct the Grounded Ophthalmic Understanding benchmark to evaluate diagnostic classification, report generation quality, and fine-grained clinical quality. Despite using only LoRA-based fine-tuning of a 7B-parameter Qwen2 backbone, RetiBridge outperforms all evaluated open-source 7B and 32B baselines, achieving the highest quantitative accuracy, evidence grounding, coverage completeness, and BERTScore, while surpassing OpenAI-o3 on these key biomarker-grounded metrics. Our code and data are released at \href{https://github.com/NatsuGao7/GROK-A-Retinal-MLLMs.git}{\texttt{RetiBridge's repository}}. 
\end{abstract}


\section{Introduction}
\label{sec:introduction}
\IEEEPARstart{R}{etinal} imaging provides a non-invasive window into both ocular and systemic health by revealing quantifiable structural and vascular features, such as retinal vessel calibre and tortuosity, optic disc morphology, and retinal layer thickness. These measurable features serve as retinal biomarkers associated with diseases ranging from diabetic retinopathy and glaucoma to hypertension and neurodegeneration \cite{wong2006retinal,schuman1995quantification}. These biomarkers are primarily derived from two complementary retinal imaging modalities. Color fundus photography (CFP) captures en-face vascular and surface features, whereas optical coherence tomography (OCT) provides depth-resolved measurements of retinal layers and macular structure \cite{keane2014retinal}. Their integration therefore provides a more comprehensive characterization of retinal pathology than either modality alone.

Traditional retinal image analysis pipelines typically combine automated segmentation of retinal structures, such as vessels, the optic disc, and retinal layers, with subsequent clinician interpretation. Representative CFP-based tools, such as AutoMorph \cite{zhou2022automorph}, can quantify biomarkers including vessel calibre and tortuosity. However, most existing methods remain modality-specific and therefore fail to fully integrate the complementary information provided by CFP and OCT. For example, in diabetic retinopathy, CFP reveals surface lesions such as microaneurysms and haemorrhages, whereas OCT captures depth-resolved structural changes, including retinal thickness and macular morphology. Moreover, although these pipelines can extract quantitative biomarkers, translating the measurements into clinically meaningful interpretations and diagnostic conclusions still relies heavily on specialist review, requiring substantial clinician time and expertise and limiting large-scale deployment \cite{holland2025specialized, haghighi2025compact}.

\begin{figure}[!t]
  \centering
  \includegraphics[width=\columnwidth]{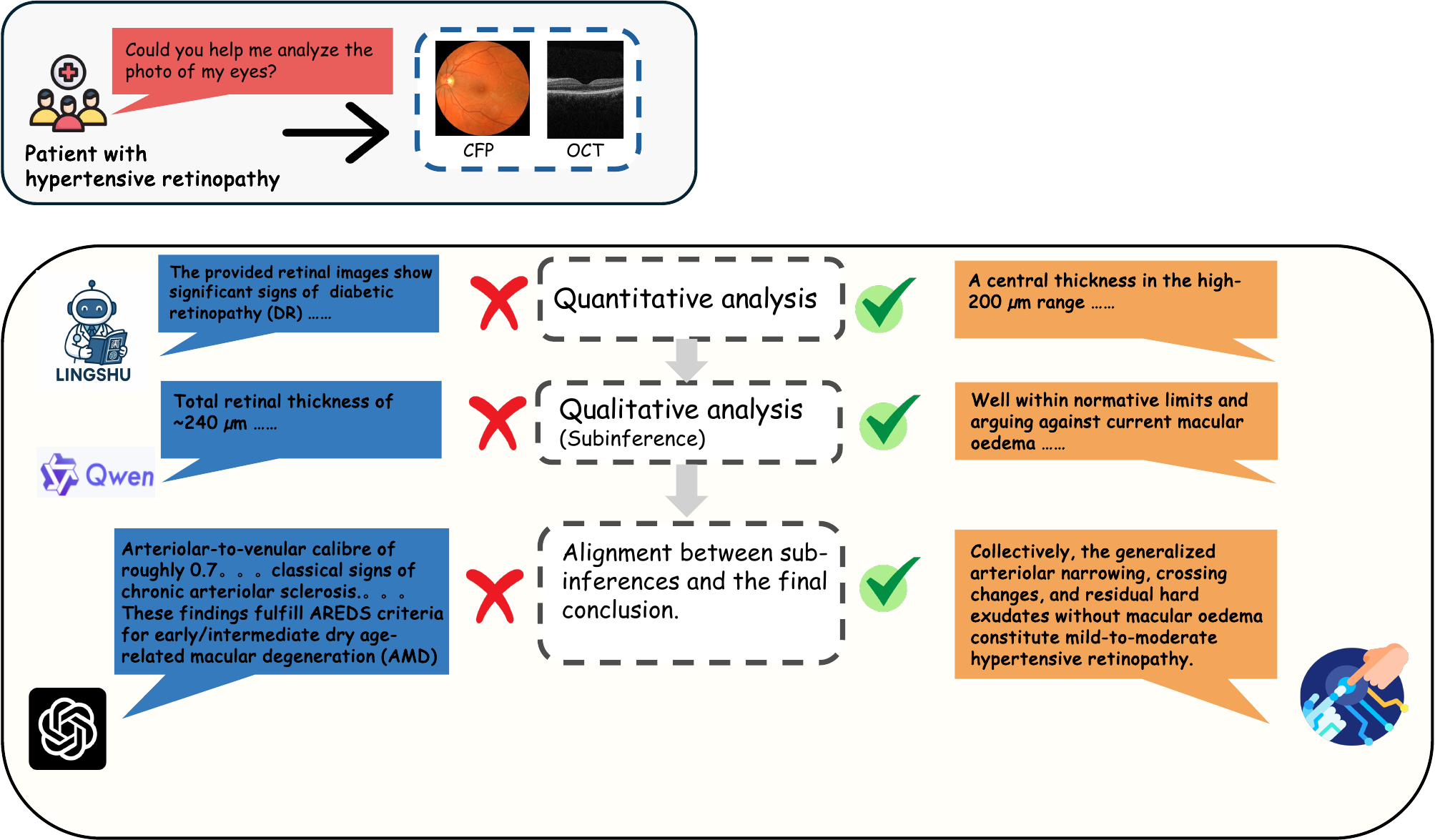}
  \caption{ Motivating example. On paired CFP \& OCT images, existing MLLMs (Lingshu-32B, Qwen2.5VL-32B, OpenAI-o3; left) each fail one of three criteria—quantitative biomarker analysis, qualitative diagnosis (sub-inference), or alignment with the final conclusion (\xmark). Our pipeline (right) meets all three (\cmark), converting measurements into coherent clinical reasoning.}
  \label{figure_1}
  \vspace{-0.2cm}
\end{figure}

To reduce reliance on manual interpretation, recent retinal foundation models, such as RETFound \cite{zhou2023foundation} and RetiZero \cite{wang2024common}, have improved automated disease prediction. However, their interpretability remains largely limited to post-hoc visual attribution maps \cite{selvaraju2017grad}, which highlight image regions associated with a prediction but do not explain how quantitative retinal biomarkers support the final diagnosis. Medical multimodal large language models (MLLMs) offer a way to address this limitation by generating natural-language interpretations of retinal findings and diagnostic decisions. General medical MLLMs, such as \textit{LLaVA-Med}
\cite{li2023llava} and \textit{Lingshu}
\cite{xu2025lingshu}, have demonstrated the potential of
generating natural-language interpretations from medical images.
Meanwhile, ophthalmology-specific models, including
\textit{OphGLM} \cite{gao2023ophglm},
\textit{RetinaVLM} \cite{holland2024specialized},
\textit{VisionUnite} \cite{li2025visionunite}, and
\textit{RetinalGPT} \cite{zhu2025retinalgpt},
have further advanced retinal image understanding,
clinical report generation, and diagnostic reasoning. Nevertheless, a critical limitation remains: \textbf{\textit{Limited clinician-friendly interpretability across modalities.}} For diagnostic settings involving paired CFP and OCT images, a
clinically useful and trustworthy ophthalmic MLLM should jointly
model their complementary information. It should be able to
\emph{quantify abnormal biomarkers} from each modality,
\emph{translate these measurements into clinically meaningful
qualitative sub-inferences}, and finally
\emph{connect these sub-inferences to a coherent diagnostic conclusion}. As \autoref{figure_1} illustrates, representative medical MLLMs can
generate textual explanations from paired CFP and OCT images, but may
fail in one or more of three key aspects:
(i) \emph{quantification of abnormal biomarkers};
(ii) translation of quantitative measurements into clinically meaningful
\emph{qualitative sub-inferences}; and
(iii) coherent linkage between these sub-inferences and the final
diagnostic conclusion.
These limitations reduce clinical transparency and reliability,
constraining the broader deployment of ophthalmic MLLMs.

To address these limitations, we introduce \textbf{RetiBridge},
a knowledge-guided ophthalmic MLLM that jointly interprets paired
CFP and OCT images, grounds its outputs in modality-specific
quantitative retinal biomarkers, translates abnormal measurements
into clinically meaningful qualitative sub-inferences, and connects
them to a coherent diagnostic conclusion. \textbf{RetiBridge} differs from existing ophthalmic MLLMs in three key respects. First, \textbf{RetiBridge} employs knowledge-guided instruction
supervision constructed from 31 OCT and 6 CFP biomarkers. These
quantitative measurements are incorporated into an expert-designed
Eye-Guideline prompt and processed by OpenAI-o3 \cite{openai2025o3} to generate
biomarker-grounded clinical interpretations. Second, RetiBridge introduces CLIP-style OCT and CFP--biomarker alignment
and integrates paired CFP and OCT representations, grounding
multimodal retinal features in clinically meaningful quantitative
measurements. Third, RetiBridge explicitly organizes diagnostic reasoning as a
quantitative-measurement-to-qualitative-sub-inference-to-diagnosis
pathway, producing a biomarker-grounded rationale that is clinically
interpretable and verifiable.

To realize these capabilities, \textbf{RetiBridge} follows a three-stage training framework. First, a doctor-verified \emph{Eye-Guideline} prompt converts 31 OCT and 6 CFP biomarkers
from 15,611 paired CFP--OCT samples into biomarker-grounded instruction responses using OpenAI-o3 \cite{openai2025o3}. Second, RetiBridge employs the pretrained CLIP-style RetiZero encoder \cite{wang2024common} for CFP and performs CLIP-style contrastive alignment between OCT B-scans and their corresponding quantitative biomarkers, addressing the lack of a comparable biomarker-grounded OCT encoder. Third,
projection modules map CFP and OCT features into the LLM embedding
space for multimodal instruction tuning and biomarker-grounded
diagnostic report generation.

The main contributions of this work are:

\textbf{\textit{Biomarker-Grounded Retinal MLLM.}} We present \textbf{RetiBridge}, a knowledge-guided ophthalmic MLLM that jointly interprets paired CFP and OCT images and explicitly bridges quantitative retinal biomarkers, clinically meaningful qualitative sub-inferences, and coherent diagnostic conclusions. This structured quantitative-to-qualitative pathway enables biomarker-grounded and clinically verifiable retinal diagnosis.

\textbf{\textit{Knowledge-Guided Instruction Data and Benchmark.}} Using 15,611 paired CFP--OCT samples from UK Biobank \cite{sudlow2015uk}, each associated with 31 OCT and 6 CFP biomarkers, we develop a reproducible Eye-Guideline pipeline for generating biomarker-grounded instruction--response data. Building on these resources, we establish the Grounded Ophthalmic Understanding benchmark to evaluate diagnostic classification, report generation quality, and fine-grained clinical quality.

\textbf{\textit{Biomarker Alignment and Cross-Modal Fusion.}}
RetiBridge adopts an asymmetric dual-encoder architecture that
combines a pretrained CLIP-style RetiZero encoder \cite{wang2024common} for CFP with a
biomarker-aligned OCT encoder. A CLIP-style objective contrastively
aligns central-foveal OCT B-scan representations with their
corresponding 31-dimensional biomarker vectors, while projection
modules map CFP and OCT features into a shared language-compatible
space for multimodal diagnostic reasoning.

\section{RELATED WORK}
\label{sec:RELATED WORK}

\subsection{MLLMs-based Ophthalmic Data Analysis}
Recent ophthalmic vision--language models have explored diverse
directions, ranging from single-modality retinal interpretation and
multimodal representation learning to lesion-aware and
knowledge-guided diagnostic reasoning. OphGLM
\cite{gao2023ophglm}, VisionUnite \cite{li2025visionunite},
and RetinalGPT \cite{zhu2025retinalgpt} primarily operate on
CFP or fundus photographs, whereas RetinaVLM
\cite{holland2024specialized} focuses on OCT-based clinical
interpretation. Notably, RetinalGPT supports quantitative vascular
analysis from CFP, but does not integrate complementary
OCT-derived biomarkers.

EyecareGPT \cite{li2025eyecaregpt} broadens ophthalmic image
understanding across heterogeneous imaging modalities. In parallel,
the non-generative vision--language model EyeCLIP
\cite{shi2025multimodal} learns shared representations across
multiple ophthalmic modalities, including CFP and OCT, for
classification, visual question answering, and cross-modal retrieval.
However, supporting multiple imaging modalities does not itself
provide patient-level paired CFP--OCT reasoning grounded in
continuous quantitative biomarkers.

Meanwhile, FundusExpert \cite{liu2025constructing},
OphthaReason \cite{wu2025bridging}, and Fundus-R1
\cite{deng2026fundus} introduce localization-aware, stepwise,
or knowledge-aware reasoning for ophthalmic diagnosis. Their
reasoning processes, however, are primarily grounded in visual
findings, image labels, and clinical knowledge rather than
continuous quantitative measurements from paired CFP and OCT.

To the best of our knowledge, existing studies have not unified
patient-level paired CFP--OCT inputs, continuous quantitative
biomarkers from both modalities, and an explicit diagnostic pathway
that connects quantitative measurements to clinically meaningful
sub-inferences and ultimately to the final diagnosis within a
report-generating ophthalmic MLLM. \textbf{RetiBridge} addresses
this gap through biomarker-grounded CFP--OCT integration and
structured quantitative-to-qualitative diagnostic reasoning.

\section{Methodology}

\begin{figure*}[!t]
  \centering
  \includegraphics[width=\textwidth]{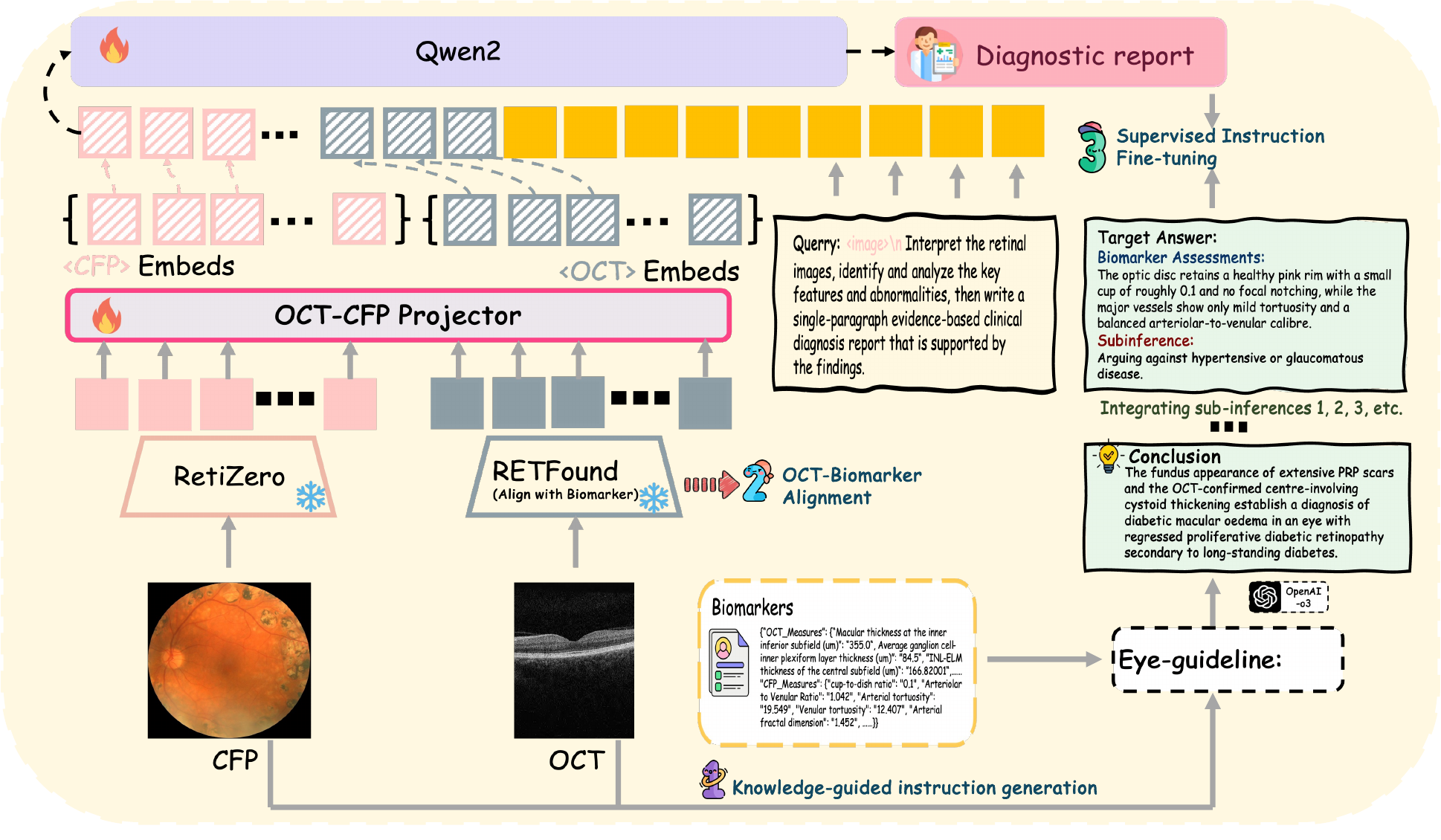} 
  \caption{Illustration of the \textbf{RetiBridge} architecture and training
workflow. RetiBridge adopts an asymmetric dual-encoder design,
using a pretrained CLIP-style RetiZero encoder \cite{wang2024common} for CFP and a
biomarker-aligned encoder for OCT. Projection modules map the CFP
and OCT representations into a shared language-compatible embedding
space for Qwen2-based diagnostic report generation. The model is
trained through three stages: Knowledge-Guided Instruction Generation,
CLIP-Style OCT--Biomarker Alignment, and Supervised Instruction
Fine-Tuning.}
  \label{figure_2}
  \vspace{-0.3cm}
\end{figure*}

\textit{Overview:} Figure \ref{figure_2} illustrates the architecture and training workflow of \textbf{RetiBridge}. Following the LLaVA paradigm\cite{liu2023visual}, RetiBridge adopts an asymmetric dual-encoder architecture: a pretrained CLIP-style RetiZero encoder \cite{wang2024common} extracts CFP representations, while a biomarker-aligned OCT encoder processes central-foveal OCT B-scans. Features from both modalities are projected into the LLM embedding space and concatenated with the textual query, enabling Qwen2 \cite{yang2024qwen2technicalreport} to generate a fine-grained diagnostic report. RetiBridge is trained through a three-stage pipeline. \textbf{Stage I: Knowledge-guided instruction generation.} We design a domain-specific prompt template, termed \textit{Eye-Guideline}, to guide OpenAI-o3 \cite{openai2025o3} in generating structured, biomarker-grounded diagnostic reports from paired CFP--OCT images, quantitative biomarkers, and diagnostic labels. The diagnostic label is used only as a soft constraint during teacher-based report generation and is not included in the multimodal input to RetiBridge during training or inference. \textbf{Stage II: CLIP-style OCT--biomarker alignment.} We contrastively align central-foveal OCT B-scan representations with their corresponding quantitative OCT biomarker vectors, thereby grounding OCT features in clinically meaningful structural measurements. For CFP, we directly leverage the pretrained CLIP-style RetiZero encoder \cite{wang2024common}, whereas the additional biomarker alignment is introduced specifically for OCT because a comparable biomarker-grounded OCT encoder is unavailable. \textbf{Stage III: Supervised instruction fine-tuning.} Using the instruction data generated in Stage I, we optimize the cross-modal projection modules together with LoRA parameters of the Qwen2 \cite{yang2024qwen2technicalreport} backbone. The projection modules map CFP and OCT features into a shared language-compatible embedding space for multimodal diagnostic report generation.

\subsection*{Preliminaries}

Let

\[
\mathcal{D}
=
\left\{
\left(
I_n^{c},
I_n^{o},
B_n^{c},
B_n^{o},
g_n
\right)
\right\}_{n=1}^{N}
\]

denote a patient-level multimodal dataset containing $N$ paired
CFP images $I_n^{c}$ and OCT B-scans $I_n^{o}$. Here,
$B_n^{c} \in \mathbb{R}^{6}$ and
$B_n^{o} \in \mathbb{R}^{31}$ denote the corresponding CFP- and
OCT-derived quantitative biomarker vectors, respectively, and $g_n$
denotes the diagnostic label of the $n$-th participant.

The CFP and OCT encoders produce token-level visual representations:

\[
Z_n^{c}=f^{c}(I_n^{c}),
\qquad
Z_n^{o}=f^{o}(I_n^{o}).
\]

For OCT--biomarker contrastive alignment, the OCT token features are
aggregated into a global representation

\[
\bar{z}_n^{o}
=
\operatorname{Pool}(Z_n^{o}),
\]

while the corresponding OCT biomarker vector is encoded as

\[
z_n^{b}
=
f^{b}(B_n^{o}),
\]

where $f^{b}(\cdot)$ is an MLP-based biomarker encoder.

\subsection*{Knowledge Guided Instruction
Generation}

We construct the knowledge-guided instruction data used to train
\textbf{RetiBridge}, as illustrated in \autoref{figure_3}(1).
The knowledge-guided component is implemented through a
clinician-reviewed prompt template, termed \textit{Eye-Guideline}.
The template incorporates ophthalmic domain knowledge and guides
OpenAI-o3 to generate structured, biomarker-grounded diagnostic
rationales. Each generated report links quantitative retinal
abnormalities to clinically meaningful qualitative interpretations
and subsequently to the final diagnostic conclusion.

For each sample in $\mathcal{D}$, let $g_n$ denote the corresponding
diagnostic label derived from UK Biobank records \cite{sudlow2015uk}. The instruction
generation process incorporates the paired CFP and OCT images,
their corresponding quantitative biomarkers, and the diagnostic label
$g_n$.

The OCT biomarkers are derived from structural measurements provided
in UK Biobank Category 100079 \cite{sudlow2015uk}, including macular thickness, volume,
area, and retinal layer thickness across multiple ETDRS subfields.
After excluding variables with more than 20\% missing values, 31 OCT
biomarkers are retained. In addition, six CFP-derived biomarkers,
including vertical cup-to-disc ratio, arteriovenous ratio, arterial and
venous fractal dimensions, and vessel tortuosity, are extracted using
the pipeline described in Section IV.

During teacher-based report generation, the diagnostic label $g_n$ is
provided to OpenAI-o3 \cite{openai2025o3} as a soft condition to improve diagnostic
consistency and constrain irrelevant generations. The generated report
$y_n$ explicitly contains the diagnosis associated with $g_n$ together
with its biomarker-grounded rationale. Therefore, $g_n$ provides
label-level supervision through the target report $y_n$, but is not
included in the multimodal input to RetiBridge during supervised
fine-tuning or inference.

For each sample in $\mathcal{D}$, the target diagnostic report is
generated as

\begin{equation}
y_n
=
\mathrm{OpenAI\text{-}o3}
\left(
\mathrm{EyeGuideline}
\left(
I_n^c,
I_n^o,
B_n^c,
B_n^o,
g_n
\right)
\right).
\end{equation}

By augmenting each sample in $\mathcal{D}$ with its generated target
report $y_n$, we define the instruction dataset as

\[
\mathcal{D}_{\mathrm{inst}}
=
\left\{
\left(
I_n^c,
I_n^o,
B_n^c,
B_n^o,
g_n,
y_n
\right)
\right\}_{n=1}^{N}.
\]

\subsection*{CLIP‑Style OCT–Biomarker Alignment
}

Although a pretrained CLIP-style encoder is available for CFP, for
which we adopt RetiZero \cite{wang2024common}, a comparable OCT
encoder explicitly grounded in continuous quantitative structural
measurements is unavailable in our setting. Moreover, conventional
image--text pretraining does not directly supervise the numerical OCT
attributes used in our instruction data. We therefore introduce a
CLIP-style OCT--biomarker alignment stage that contrastively aligns
central-foveal OCT B-scan representations with their corresponding
31-dimensional OCT biomarker vectors derived from volumetric
measurements. We use the central-foveal B-scan because it provides a
standardized anatomical cross-section through the fovea and directly
captures key macular and retinal-layer structures represented by the
retained OCT biomarkers.

Given a mini-batch of $M_{\mathrm{align}}$ matched OCT--biomarker
pairs, the OCT encoder $f^{o}(\cdot)$ first produces token-level
representations

\[
Z_j^{o}=f^{o}(I_j^{o}),
\]

which are aggregated into global OCT representations

\[
\bar{z}_j^{o}
=
\operatorname{Pool}(Z_j^{o}).
\]

Meanwhile, the MLP-based biomarker encoder $f^{b}(\cdot)$ maps the
corresponding OCT biomarker vectors to

\[
z_j^{b}
=
f^{b}(B_j^{o}).
\]

As illustrated in \autoref{figure_3}(2), matched OCT--biomarker
representations are encouraged to have high cosine similarity, whereas
unmatched pairs within the mini-batch are pushed apart. The
OCT-to-biomarker loss is defined as

\begin{equation}
\ell_{o \rightarrow b}
=
-\frac{1}{M_{\mathrm{align}}}
\sum_{j=1}^{M_{\mathrm{align}}}
\log
\frac{
\exp\left(
\operatorname{sim}(\bar{z}_j^{o},z_j^{b})/\tau
\right)
}{
\sum_{k=1}^{M_{\mathrm{align}}}
\exp\left(
\operatorname{sim}(\bar{z}_j^{o},z_k^{b})/\tau
\right)
},
\end{equation}

where $\operatorname{sim}(\cdot,\cdot)$ denotes cosine similarity and
$\tau$ is the temperature hyperparameter. The symmetric
biomarker-to-OCT loss is defined as

\begin{equation}
\ell_{b \rightarrow o}
=
-\frac{1}{M_{\mathrm{align}}}
\sum_{j=1}^{M_{\mathrm{align}}}
\log
\frac{
\exp\left(
\operatorname{sim}(z_j^{b},\bar{z}_j^{o})/\tau
\right)
}{
\sum_{k=1}^{M_{\mathrm{align}}}
\exp\left(
\operatorname{sim}(z_j^{b},\bar{z}_k^{o})/\tau
\right)
}.
\end{equation}

The final alignment objective is

\begin{equation}
\mathcal{L}_{\mathrm{align}}
=
\frac{1}{2}
\left(
\ell_{o \rightarrow b}
+
\ell_{b \rightarrow o}
\right).
\end{equation}

After alignment, the biomarker-aligned OCT encoder is retained to
extract token-level OCT representations for downstream multimodal
instruction tuning. The OCT biomarker vectors are used only during
the alignment stage and are not required during RetiBridge inference.

\subsection*{Supervised Instruction Fine-Tuning}

The CFP and OCT encoders produce modality-specific token-level
representations with different feature dimensions and distributions,
neither of which is directly mapped to the token embedding space of
the language model. We therefore introduce projection modules to map
both visual streams into a shared language-compatible representation.

\begin{figure*}[!t]
  \centering
  \includegraphics[width=\textwidth]{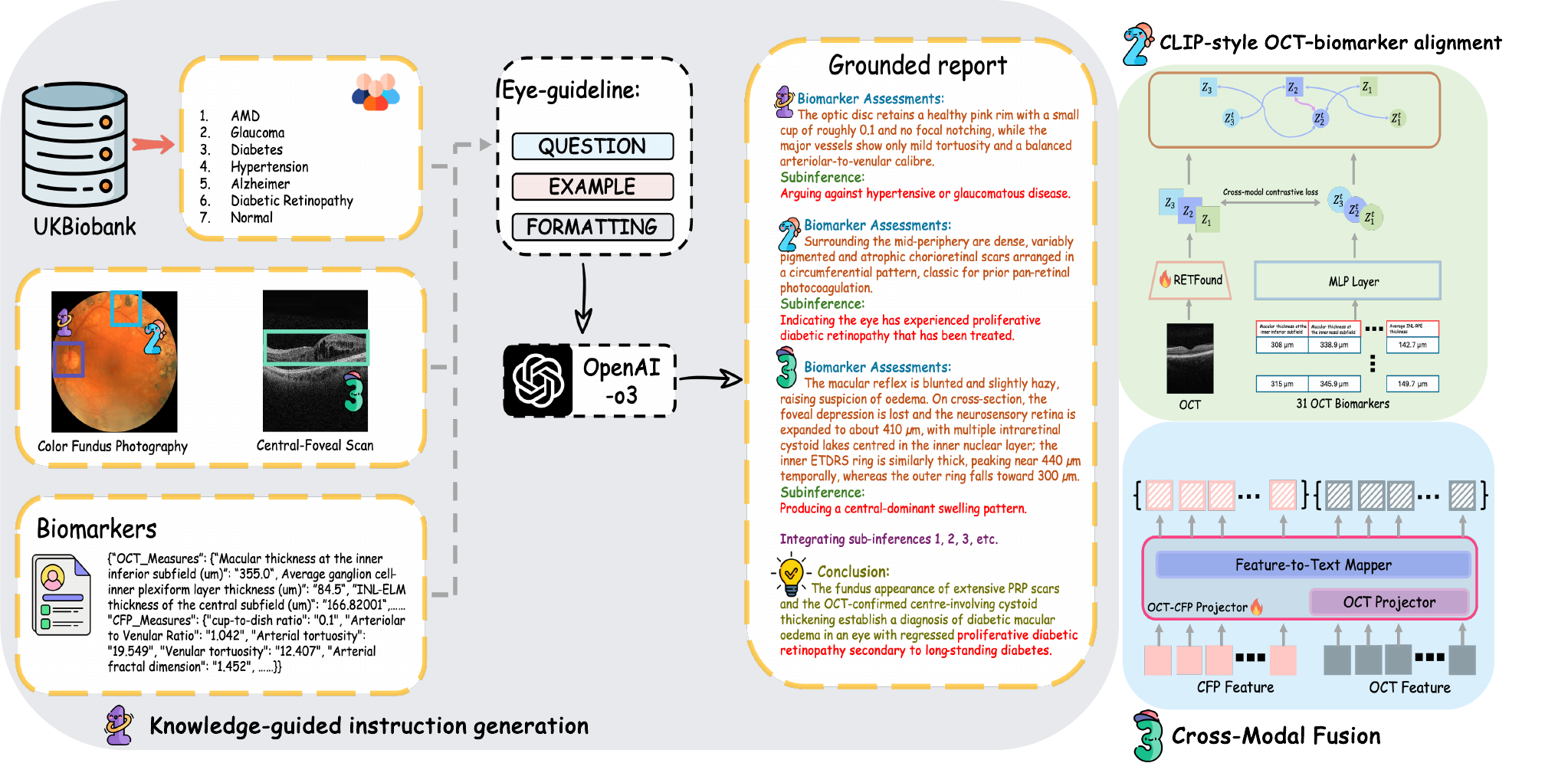}
  \caption{
  Eye-Guideline and OpenAI-o3 generate structured diagnostic reports
grounded in paired CFP/OCT images, quantitative biomarkers, and
diagnostic labels.
\textbf{CLIP-Style OCT--Biomarker Alignment:}
Central-foveal OCT B-scan representations are contrastively aligned
with corresponding 31-dimensional OCT biomarker vectors.
\textbf{Cross-Modal Fusion and Instruction Fine-Tuning:}
Projected CFP and OCT features are integrated in the Qwen2 embedding
space for biomarker-grounded diagnostic report generation.}
  \label{figure_3}
\end{figure*}

As illustrated in \autoref{figure_3}(3), the pretrained OCT and CFP
encoders, denoted by $f^{o}(\cdot)$ and $f^{c}(\cdot)$, respectively,
are frozen during supervised instruction fine-tuning. Given a
mini-batch of $M_{\mathrm{SFT}}$ paired retinal images, the two
encoders produce token-level visual representations

\begin{equation}
\begin{split}
Z^{o} &= f^{o}(I^{o}) \in \mathbb{R}^{M_{\mathrm{SFT}} \times L_{o} \times d_{o}}, \\
Z^{c} &= f^{c}(I^{c}) \in \mathbb{R}^{M_{\mathrm{SFT}} \times L_{c} \times d_{c}}.
\end{split}
\end{equation}
where $M_{\mathrm{SFT}}$ denotes the mini-batch size during supervised
instruction fine-tuning, $L_{o}$ and $L_{c}$ denote the numbers of OCT
and CFP visual tokens, and $d_{o}$ and $d_{c}$ denote their respective
feature dimensions. Unlike the pooled OCT representation
$\bar{z}_n^{o}$ used for contrastive alignment in Stage II, $Z^{o}$
contains token-level OCT features used for multimodal fusion.

Because the OCT and CFP encoders produce representations with
different feature dimensions, an OCT-specific MLP projector first maps
the OCT features to the dimensionality of the CFP features:

\begin{equation}
\hat{Z}^{o}
=
\mathrm{MLP}_{o}
\left(
Z^{o};\theta_{P_o}
\right)
\in
\mathbb{R}^{M_{\mathrm{SFT}} \times L_{o} \times d_{c}},
\end{equation}

where $\theta_{P_o}$ denotes the parameters of the OCT-specific
projector. A shared feature-to-text projector then maps the OCT and CFP
representations into the $d_t$-dimensional token embedding space of
Qwen2:

\begin{equation}
H^{o}
=
\mathrm{MLP}_{\mathrm{text}}
\left(
\hat{Z}^{o};\theta_{P_t}
\right)
\in
\mathbb{R}^{M_{\mathrm{SFT}} \times L_{o} \times d_{t}},
\end{equation}

\begin{equation}
H^{c}
=
\mathrm{MLP}_{\mathrm{text}}
\left(
Z^{c};\theta_{P_t}
\right)
\in
\mathbb{R}^{M_{\mathrm{SFT}} \times L_{c} \times d_{t}},
\end{equation}

where $\theta_{P_t}$ denotes the parameters of the shared
feature-to-text projector and $d_t$ is the dimensionality of the Qwen2
token embedding space.

For the $n$-th sample, the projected CFP and OCT tokens are concatenated
with the embedded textual query $x_{q,n}$ along the token dimension to
construct the multimodal input

\begin{equation}
x_n
=
\mathrm{Concat}
\left(
H_n^{c},
H_n^{o},
\mathrm{Embed}(x_{q,n})
\right).
\end{equation}

Here, $x_{q,n}$ is a standardized instruction asking the model to
interpret the paired retinal images and generate a biomarker-grounded
diagnostic report.

During supervised instruction fine-tuning, the pretrained RetiZero CFP
encoder \cite{wang2024common} and the biomarker-aligned RETFound OCT
encoder \cite{zhou2023foundation} remain frozen. We optimize the
OCT-specific projector, the shared feature-to-text projector, and the
LoRA parameters of the Qwen2-7B-Instruct backbone using the synthetic
instruction targets generated in Stage I.

For each sample, OpenAI-o3 \cite{openai2025o3} uses the diagnostic label $g_n$ as a soft
condition to generate the target report $y_n$ in Stage I. The generated
report explicitly contains the diagnosis associated with $g_n$ together
with its biomarker-grounded rationale. Therefore, $g_n$ provides
label-level supervision to RetiBridge through the target report $y_n$,
but is never included as an input feature in $x_n$ during training or
inference.

Similarly, the quantitative biomarker vectors are not directly included
in $x_n$: both CFP and OCT biomarkers are used to construct the target
report $y_n$ in Stage I, while the OCT biomarkers are additionally used
for OCT--biomarker alignment in Stage II.

Let

\[
\Theta
=
\left\{
\theta_{P_o},
\theta_{P_t},
\theta_{\mathrm{LoRA}}
\right\}
\]

denote the set of trainable parameters. Given the target report
$y_n=(y_{n,1},\ldots,y_{n,T_n})$, the supervised instruction
fine-tuning objective is

\begin{equation}
\mathcal{L}_{\mathrm{SFT}}
=
-
\frac{1}{M_{\mathrm{SFT}}}
\sum_{n=1}^{M_{\mathrm{SFT}}}
\sum_{i=1}^{T_n}
\log
P_{\Theta}
\left(
y_{n,i}
\mid
x_n,y_{n,<i}
\right),
\end{equation}

where $T_n$ denotes the length of the $n$-th target report and
$y_{n,<i}$ denotes all target tokens preceding $y_{n,i}$. The loss is
computed only over the target report tokens, while the multimodal input
and instruction tokens serve as conditioning context.

\section{EXPERIMENTS AND RESULTS}
\subsection*{Dataset}
We construct a patient-level multimodal dataset from UK Biobank:

\[
\mathcal{D}
=
\left\{
\left(
I_n^{c},
I_n^{o},
B_n^{c},
B_n^{o},
g_n
\right)
\right\}_{n=1}^{N},
\]

where $I_n^{c}$ and $I_n^{o}$ denote the paired CFP and OCT images
of the $n$-th participant, $B_n^{c}$ and $B_n^{o}$ denote the
corresponding CFP- and OCT-derived biomarker vectors, and $g_n$
denotes the diagnostic label.

For each participant, identified by the unique UK Biobank \cite{sudlow2015uk} identifier
\texttt{eid}, we extract the left-eye color fundus photograph from
Field 21015 and the corresponding central-foveal OCT B-scan from
Field 21017. Because the required CFP-derived quantitative biomarkers
are not directly provided by UK Biobank \cite{sudlow2015uk}, we adopt an EyeQ-inspired
preprocessing pipeline \cite{fu2019evaluation}. Specifically, CFP
images are first quality-filtered using a ResNet-based model
\cite{he2016deep} trained on EyePACS. Images passing quality control
are then processed using nnUNetv2 \cite{isensee2021nnu} for
artery--vein segmentation and BI-GCN \cite{meng2021graph} for
optic cup and disc segmentation.

From the resulting segmentation maps, we extract six CFP-derived
biomarkers, denoted by $B_n^{c}\in\mathbb{R}^{6}$. We additionally
retain 31 OCT-derived structural biomarkers from UK Biobank \cite{sudlow2015uk},
denoted by $B_n^{o}\in\mathbb{R}^{31}$. Therefore, each participant
is associated with a total of 37 quantitative retinal biomarkers.

To derive the diagnostic label $g_n$, we integrate multiple UK
Biobank data sources, including self-reported non-cancer illness
codes from Field 20002, hospital episode statistics with ICD-10
diagnoses, and relevant clinical fields. Based on predefined
phenotype rules, we select six representative ophthalmic and systemic
disease categories for analysis, as summarized in
\autoref{tab:data_distribution}.

The dataset is split at the participant level according to
\texttt{eid}, ensuring that no participant appears in both subsets.
An 80:20 split yields 12,489 training samples and 3,122 testing
samples.

\begin{table}[htbp]
\centering
\scriptsize
\setlength{\tabcolsep}{3.5pt} 
\caption{Distribution (\%) of disease categories in the dataset.}
\label{tab:data_distribution}
\begin{tabular}{ccccccc}
\toprule
Normal & Hypertension & Diabetes & Glaucoma & DR & AMD & Alzheimer \\
\midrule
38.40 & 36.05 & 19.95 & 3.40 & 1.16 & 0.86 & 0.18 \\
\bottomrule
\label{tab:data_distribution}
\end{tabular}
\vspace{-0.3cm}
\begin{tablenotes}
\item[*] The “Normal” group does not indicate complete health, but rather the absence of the six listed ophthalmic and systemic diseases.
\end{tablenotes}
\end{table}

\subsection*{Experimental Setting}
 In OCT-Biomarker alignment stage, each 244$\times$244 central-foveal OCT B-scan is contrastively aligned with its corresponding 3D biomarker embedding using InfoNCE loss ($\tau=0.5$) and the AdamW optimizer (learning rate = 1e-4, weight decay = 1e-2), for a total of 50 epochs. 
As for the Supervised Instruction Finetuning stage, we jointly fine-tune the LLM and its projector module. To reduce computational cost, we adopt the Low-Rank Adaptation (LoRA) technique. Training is conducted on a single NVIDIA L40S GPU (48\,GB VRAM) for 2 epochs.

Note that both stages use the same \texttt{eid}-based split of the UKBiobank dataset to define training and testing sets, ensuring consistency across the two phases.

\subsection*{Baseline Model}
We compare \textbf{RetiBridge} with six representative multimodal large language models, including the proprietary models GPT-4o \cite{openai2024gpt4o} and OpenAI-o3 \cite{openai2025o3}, the general-purpose open-source models Qwen2.5-VL-7B \cite{bai2025qwen25vltechnicalreport} and Qwen2.5-VL-32B \cite{bai2025qwen25vltechnicalreport}, and the medically oriented models Lingshu-7B \cite{xu2025lingshu} and Lingshu-32B \cite{xu2025lingshu}. These models can be evaluated using the same paired CFP--OCT inputs, diagnostic prompt, and report-level metrics as RetiBridge, enabling a consistent comparison across different model scales and levels of medical specialization. Ophthalmology-specific models discussed in the related-work section are not included in the quantitative comparison because their supported modalities, task settings, and output formats differ from our paired CFP--OCT diagnostic report-generation setting. 

We additionally evaluate four controlled variants of RetiBridge. \textbf{RetiBridge-CLIP-ViT} replaces the proposed asymmetric retinal encoders with a vanilla CLIP-ViT jointly trained on CFP and OCT. \textbf{RetiBridge-RETFound-OCT} replaces the biomarker-aligned OCT encoder with the original RETFound encoder \cite{zhou2023foundation}. \textbf{RetiBridge-Vicuna-7B} replaces Qwen2-7B-Instruct with Vicuna-7B-v1.5, while \textbf{RetiBridge-OCT-only} removes the CFP branch. These variants are used to assess the contributions of the biomarker-aligned OCT encoder, the language-model backbone, and paired CFP--OCT integration.

\subsection*{Grounded Ophthalmic Understanding Benchmark}

\begin{figure*}[!t]
  \centering
   \includegraphics[width=\textwidth]{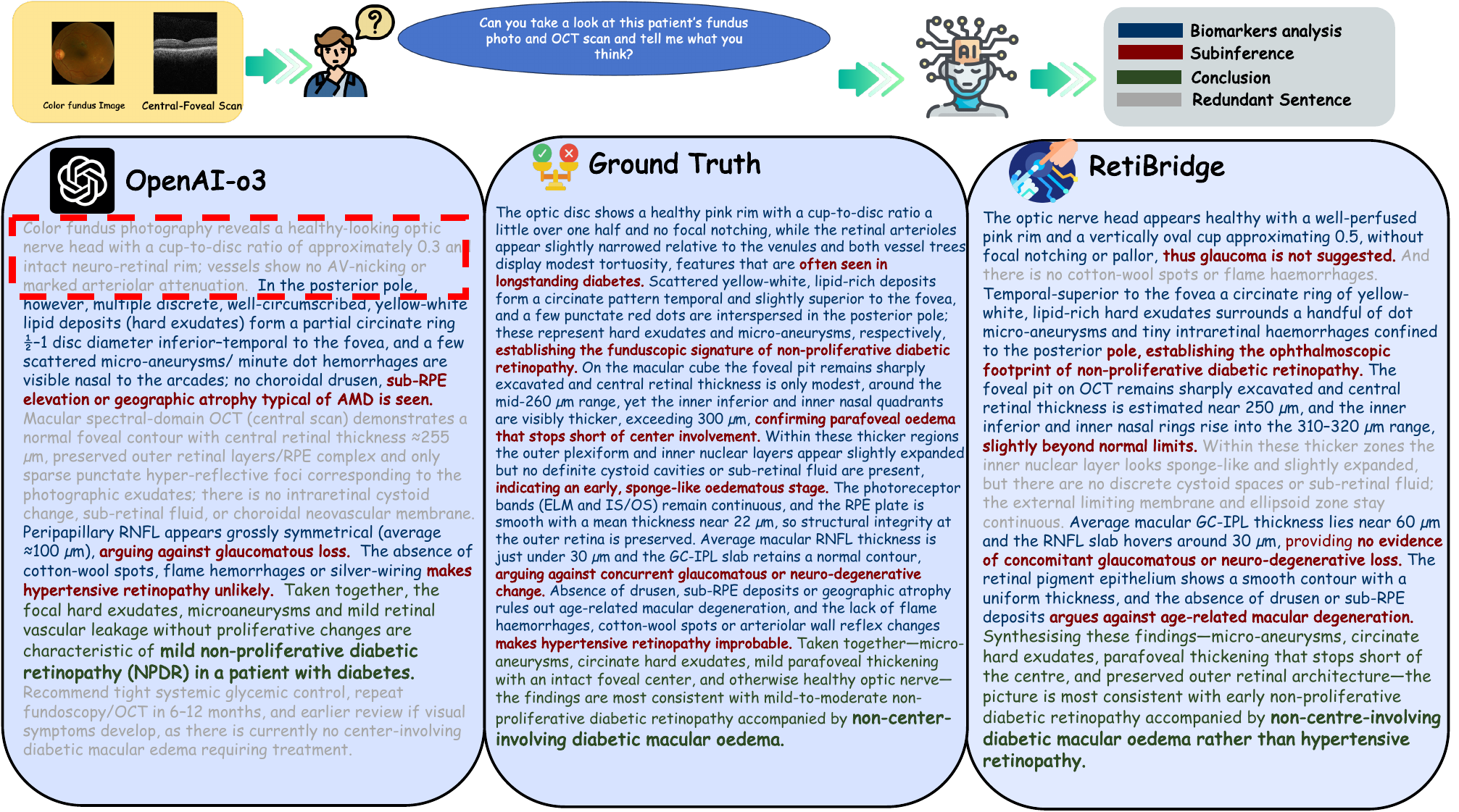}
  \caption{Qualitative comparison of biomarker-grounded diagnostic reports generated
by \textbf{RetiBridge} and OpenAI-o3 for a diabetic retinopathy case
using paired CFP and OCT images.}
  \label{figure_4}
\end{figure*}

We establish the \textbf{Grounded Ophthalmic Understanding benchmark}
to evaluate structured diagnostic report generation from paired CFP
and OCT images, as illustrated in \autoref{figure_4}. The benchmark
combines diagnostic labels, teacher-generated reference reports,
quantitative retinal biomarkers, and a unified evaluation protocol.
Model performance is assessed from three complementary perspectives:

\begin{itemize}

    \item \textbf{Diagnostic Classification Performance:}
    We extract the final predicted diagnosis from each generated report
    and compare it with the corresponding diagnostic label $g_n$.
    Macro F1-score is used to account for class imbalance across the
    evaluated disease categories.

    \item \textbf{Report-Level Semantic Similarity:}
    We compute BERTScore between each generated report and its
    corresponding teacher-generated reference report $y_n$.
    BERTScore measures semantic similarity but does not independently
    assess clinical correctness.

    \item \textbf{Fine-Grained Clinical Quality Assessment:}
    We develop a rubric-based scoring framework to assess six
    dimensions of clinical report quality. \textit{Quantitative
    Accuracy} evaluates the correctness of numerical measurements;
    \textit{Qualitative Accuracy} evaluates the correctness of
    descriptive clinical findings; \textit{Evidence Grounding}
    assesses whether diagnostic claims are supported by the available
    images and biomarkers; \textit{Reasoning Consistency} evaluates
    whether the diagnostic rationale is coherent and non-contradictory;
    \textit{Coverage Completeness} assesses whether the report covers
    key retinal structures and findings; and \textit{Error Severity
    Score} evaluates the presence of clinically consequential or
    management-changing errors, with higher scores indicating fewer
    severe errors.

\end{itemize}

The rubric-based evaluation is implemented using Claude Opus 4.8
\cite{anthropic2026opus} as an independent LLM-based evaluator.
Claude is not used for instruction-data generation or model training
and is not included as a competing baseline. For each case, the
evaluator receives the paired CFP and OCT images, the anonymized
candidate report, the diagnostic label $g_n$, the corresponding
quantitative CFP and OCT biomarkers, and the predefined scoring
rubric. Model identities are concealed during scoring to reduce
evaluator bias.

All baseline models described in the previous subsection are evaluated
using the same paired CFP--OCT inputs, standardized diagnostic
instruction, and evaluation metrics. The standardized instruction
requests both quantitative observations and evidence-grounded
diagnostic interpretations, thereby reducing performance differences
caused solely by variations in output format.

\subsection*{Experimental Results}
\begin{table*}[t!]
  \begin{threeparttable}
  \caption{Comprehensive performance comparison on the Grounded
  Ophthalmic Understanding benchmark. All metrics except BERTScore and
  Macro F1 are reported on a 0--100 scale, with higher values indicating
  better performance. Bold and underlined values denote the best and
  second-best results among the main models, respectively.}
  \label{tab:comprehensive_comparison}
  \centering

  \begin{tabularx}{\textwidth}{@{} l *{8}{C} @{}}
    \toprule
    \textbf{Model}
    & \textbf{Quantitative}
    & \textbf{Qualitative}
    & \textbf{Evidence}
    & \textbf{Reasoning}
    & \textbf{Coverage}
    & \textbf{Error}
    & \textbf{BERT}
    & \textbf{Macro} \\

    & \textbf{Accuracy}
    & \textbf{Accuracy}
    & \textbf{Grounding}
    & \textbf{Consistency}
    & \textbf{Completeness}
    & \textbf{Severity Score}
    & \textbf{Score}
    & \textbf{F1} \\

    \midrule

    GPT-4o \cite{openai2024gpt4o}
    & 53.50
    & 40.14
    & 33.25
    & \underline{66.75}
    & 49.50
    & 57.67
    & 0.85
    & \underline{0.24} \\

    OpenAI-o3 \cite{openai2025o3}
    & \underline{58.31}
    & \textbf{47.79}
    & \underline{60.13}
    & \textbf{70.97}
    & \underline{62.70}
    & \textbf{61.60}
    & \underline{0.87}
    & \textbf{0.25} \\

    \addlinespace

    Qwen2.5-VL-32B \cite{bai2025qwen25vltechnicalreport}
    & 46.03
    & 33.25
    & 37.22
    & 65.68
    & 44.85
    & 53.08
    & 0.83
    & 0.19 \\

    Lingshu-32B \cite{xu2025lingshu}
    & 15.60
    & 35.19
    & 34.18
    & 66.61
    & 45.49
    & 51.76
    & 0.85
    & 0.23 \\

    \addlinespace

    Qwen2.5-VL-7B \cite{bai2025qwen25vltechnicalreport}
    & 29.47
    & 9.90
    & 28.14
    & 50.60
    & 21.69
    & 20.82
    & 0.83
    & 0.19 \\

    Lingshu-7B \cite{xu2025lingshu}
    & 26.84
    & 31.11
    & 26.84
    & 61.07
    & 36.08
    & 44.02
    & 0.85
    & 0.18 \\

    \addlinespace

    \textbf{RetiBridge (Ours)}
    & \textbf{78.23}
    & \underline{45.98}
    & \textbf{68.70}
    & 66.42
    & \textbf{65.56}
    & \underline{60.59}
    & \textbf{0.88}
    & 0.23 \\

    \midrule

    RetiBridge-CLIP-ViT
    & 61.52
    & 40.81
    & 65.74
    & 56.37
    & 57.68
    & 50.43
    & 0.86
    & 0.19 \\

    RetiBridge-RETFound-OCT
    & 58.79
    & 43.71
    & 63.85
    & 61.17
    & 58.12
    & 60.13
    & 0.88
    & 0.20 \\

    RetiBridge-Vicuna-7B
    & 54.23
    & 38.96
    & 57.61
    & 55.38
    & 58.21
    & 57.54
    & 0.87
    & 0.19 \\

    RetiBridge-OCT-only
    & 41.13
    & 35.17
    & 36.31
    & 48.20
    & 47.65
    & 52.01
    & 0.87
    & 0.14 \\

    \bottomrule
  \end{tabularx}

  \begin{tablenotes}
    \footnotesize

    \item \textbf{RetiBridge (Ours)} uses a pretrained RetiZero CFP
    encoder, a biomarker-aligned RETFound OCT encoder, and
    Qwen2-7B-Instruct.

    \textbf{RetiBridge-CLIP-ViT} replaces the asymmetric retinal
    encoders with a vanilla CLIP-ViT encoder jointly trained on CFP and
    OCT.

    \textbf{RetiBridge-RETFound-OCT} replaces the biomarker-aligned OCT
    encoder with the original RETFound encoder.

    \textbf{RetiBridge-Vicuna-7B} replaces Qwen2-7B-Instruct with
    Vicuna-7B-v1.5.

    \textbf{RetiBridge-OCT-only} removes the CFP branch and retains only
    the OCT encoder and Qwen2-7B-Instruct.
  \end{tablenotes}

  \end{threeparttable}
\end{table*}

The comprehensive performance of all evaluated models is summarized in
\autoref{tab:comprehensive_comparison}.

\textbf{Diagnostic classification performance.}
All evaluated models achieve relatively low Macro F1-scores
($\leq 0.25$), highlighting the difficulty of diagnostic classification
under the long-tailed class distribution shown in
\autoref{tab:data_distribution}. Despite using a 7B-parameter language
backbone, \textbf{RetiBridge} achieves a Macro F1-score of 0.23,
matching the medically oriented Lingshu-32B model and approaching
OpenAI-o3 \cite{openai2025o3}, which obtains the highest score of 0.25. RetiBridge also
outperforms the other evaluated open-source 7B models, including
Qwen2.5-VL-7B and Lingshu-7B, which achieve Macro F1-scores of 0.19
and 0.18, respectively. These results indicate that parameter-efficient
LoRA-based adaptation, together with domain-specific retinal
representations, enables RetiBridge to remain competitive with
substantially larger models.

\textbf{Report-level semantic similarity.}
RetiBridge achieves the highest BERTScore of 0.88, compared with 0.87
for OpenAI-o3 \cite{openai2025o3} and 0.85 for GPT-4o \cite{openai2024gpt4o}. This result indicates stronger
semantic similarity to the teacher-generated reference reports.
However, because these reference reports are generated using
OpenAI-o3 \cite{openai2025o3} and RetiBridge is trained on reports produced through the
same instruction-generation pipeline, BERTScore may partly reflect
similarities in report structure and linguistic style rather than
clinical correctness alone. We therefore interpret BERTScore together
with the diagnostic and fine-grained clinical metrics.

\textbf{Fine-grained clinical quality.}
RetiBridge achieves the highest scores among all evaluated models in
\textit{Quantitative Accuracy} (78.23), \textit{Evidence Grounding}
(68.70), and \textit{Coverage Completeness} (65.56). These results
suggest that the model is effective at estimating quantitative retinal
findings, linking diagnostic statements to the available clinical
evidence, and covering relevant retinal structures. RetiBridge also
achieves the second-best \textit{Qualitative Accuracy} score of 45.98
and a competitive \textit{Error Severity Score} of 60.59. OpenAI-o3 \cite{openai2025o3}
performs better in \textit{Qualitative Accuracy} (47.79) and
\textit{Reasoning Consistency} (70.97), whereas RetiBridge obtains
higher quantitative accuracy, evidence grounding, and coverage
completeness. Compared with all evaluated open-source 7B and 32B
baselines, RetiBridge demonstrates consistently stronger performance
across most grounding-related dimensions. These improvements support
the effectiveness of OCT--biomarker alignment, paired CFP--OCT
integration, and knowledge-guided instruction fine-tuning.

\textbf{Qualitative case analysis.}
As illustrated in \autoref{figure_4}, RetiBridge organizes its report
along a structured quantitative-to-qualitative diagnostic pathway. It
first estimates relevant retinal measurements from the paired CFP and
OCT images, translates abnormal measurements into clinically meaningful
sub-inferences, and then connects these findings to the final diagnostic
conclusion. In the illustrated diabetic retinopathy case, OpenAI-o3 \cite{openai2025o3}
includes several repetitive statements, highlighted by the red boxes,
that are less explicitly connected to intermediate clinical
interpretations or the final diagnosis. By contrast, RetiBridge provides
a more concise and traceable linkage between quantitative findings,
qualitative interpretations, and the diagnostic conclusion. This
example qualitatively complements the improvements observed in
quantitative accuracy, evidence grounding, and coverage completeness.

\subsection*{Ablation Study}\label{formats}

\begin{table}[h]
\centering
\caption{Comparison of OCT biomarker regression performance between
RETFound and the proposed biomarker-aligned OCT encoder. Lower MAE and
RMSE and higher $R^2$ indicate better performance.}
\label{tab:regression-results}
\begin{tabular}{lccc}
\toprule
\textbf{Model} & \textbf{MAE} & \textbf{RMSE} & $\boldsymbol{R^2}$ \\
\midrule
RETFound \cite{zhou2023foundation} & 3.8525 & 6.3583 & 0.6621 \\
Ours & \textbf{3.6069} & \textbf{6.1004} & \textbf{0.7106} \\
\bottomrule
\end{tabular}
\vspace{-0.4cm}
\end{table}

Although Macro F1 and several fine-grained clinical quality scores
remain modest, the ablation experiments demonstrate the contributions
of RetiBridge's principal architectural and training components.

\textbf{Asymmetric modality-specific retinal encoders improve
diagnostic report generation.}
As shown in \autoref{tab:comprehensive_comparison}, replacing the
combination of the pretrained CLIP-style RetiZero CFP encoder \cite{wang2024common} and the
biomarker-aligned OCT encoder with a vanilla CLIP-ViT encoder jointly
trained on CFP and OCT
(\textit{RetiBridge-CLIP-ViT}) reduces Quantitative Accuracy from
78.23 to 61.52, Qualitative Accuracy from 45.98 to 40.81, and Macro F1
from 0.23 to 0.19. These results support the benefit of using
modality-specific retinal encoders rather than a shared generic
encoder. Specifically, RetiZero \cite{wang2024common} provides semantically aligned CFP
representations, whereas the OCT encoder is explicitly grounded in
continuous quantitative structural biomarkers.

\textbf{OCT--biomarker alignment improves quantitative
representation learning.}
Replacing the biomarker-aligned OCT encoder with the original
RETFound encoder
(\textit{RetiBridge-RETFound-OCT}) reduces Quantitative Accuracy from
78.23 to 58.79. This result suggests that general OCT pretraining is
less effective than explicit OCT--biomarker alignment for the
quantitative interpretation required by our task. To further examine
this effect, we fine-tune the biomarker-aligned OCT encoder and the
original RETFound encoder \cite{zhou2023foundation} on a regression task involving the 31
OCT-derived biomarkers. As shown in
\autoref{tab:regression-results}, the biomarker-aligned encoder achieves
a lower MAE (3.6069 vs. 3.8525), a lower RMSE
(6.1004 vs. 6.3583), and a higher $R^2$
 (0.7106 vs. 0.6621). These results further support the effectiveness
of contrastive OCT--biomarker alignment for learning quantitatively
informative OCT representations.

\textbf{Paired CFP--OCT inputs provide complementary diagnostic
information.}
Removing the CFP branch
(\textit{RetiBridge-OCT-only}) decreases Quantitative Accuracy from
78.23 to 41.13, Qualitative Accuracy from 45.98 to 35.17, Evidence
Grounding from 68.70 to 36.31, Reasoning Consistency from 66.42 to
48.20, Coverage Completeness from 65.56 to 47.65, and Error Severity
Score from 60.59 to 52.01. These results demonstrate that CFP provides
substantial complementary information beyond the depth-resolved
structural features captured by OCT. In particular, CFP contributes
en-face vascular and surface information that supports more complete
and better-grounded diagnostic reports when combined with OCT.

\textbf{The language-model backbone affects multimodal diagnostic
generation.}
Replacing Qwen2-7B-Instruct with Vicuna-7B-v1.5
(\textit{RetiBridge-Vicuna-7B}) results in lower performance across
the evaluated clinical dimensions. Quantitative Accuracy decreases
from 78.23 to 54.23, Qualitative Accuracy from 45.98 to 38.96,
Evidence Grounding from 68.70 to 57.61, and Reasoning Consistency
from 66.42 to 55.38. These findings indicate that the language-model
backbone plays an important role in integrating CFP and OCT
representations and transforming multimodal evidence into structured
diagnostic reports. Investigating newer and stronger language
backbones within the RetiBridge framework is therefore a promising
direction for future work.

\section{Conclusion}
In this paper, we introduce \textbf{RetiBridge}, a knowledge-guided
ophthalmic multimodal large language model that jointly interprets
paired CFP and OCT images and generates structured diagnostic reports.
RetiBridge combines a pretrained CLIP-style RetiZero CFP encoder \cite{wang2024common} with
a biomarker-aligned RETFound OCT encoder \cite{zhou2023foundation}, and maps their complementary
representations into the language-model embedding space for
LoRA-based instruction fine-tuning. Through knowledge-guided
instruction generation and explicit OCT--biomarker alignment, the
model organizes retinal interpretation along a structured
quantitative-measurement-to-qualitative-sub-inference-to-diagnosis
pathway.

On the Grounded Ophthalmic Understanding benchmark, RetiBridge achieves
competitive diagnostic classification performance and consistently
outperforms the evaluated open-source 7B and 32B baselines across most
report-generation and fine-grained clinical quality metrics. It also
surpasses OpenAI-o3 \cite{openai2025o3} in Quantitative Accuracy, Evidence Grounding,
Coverage Completeness, and BERTScore, while OpenAI-o3 \cite{openai2025o3} retains stronger
Qualitative Accuracy and Reasoning Consistency. The ablation results
further demonstrate the contributions of modality-specific retinal
encoders, OCT--biomarker alignment, paired CFP--OCT integration, and
the language-model backbone.

These findings suggest that grounding multimodal retinal representations
in continuous quantitative biomarkers can improve the clinical
traceability and completeness of generated diagnostic reports. Future
work will extend RetiBridge to full volumetric OCT and additional
ophthalmic imaging modalities, validate its generalizability on
external clinical datasets, and investigate stronger language
backbones and broader expert evaluation.

\section*{References}
\bibliographystyle{IEEEtran}
\bibliography{reference}

\end{document}